# Terahertz Security Image Quality Assessment by No-reference Model Observers

Menghan Hu[a], Xiongkuo Min[a], Guangtao Zhai[a*], Wenhan Zhu[a], Yucheng Zhu[a], Zhaodi Wang[a], Xiaokang Yang[a], Guang Tian[b]

[a] *Institute of Image Communication and Information Processing, Shanghai Jiao Tong University, Shanghai 200240, PR China;*

[b] *BOCOM Smart Network Technologies Inc., Shanghai 200433, PR China*

* zhaiguangtao@sjtu.edu.cn;

**Abstract.**

To provide the possibility of developing objective image quality assessment (IQA) algorithms for THz security images, we constructed the THz security image database (THSID) including a total of 181 THz security images with the resolution of 127×380. The main distortion types in THz security images were first analyzed for the design of subjective evaluation criteria to acquire the mean opinion scores. Subsequently, the existing no-reference IQA algorithms, which were 5 opinion-aware approaches viz., NFERM, GMLF, DIIVINE, BRISQUE and BLIINDS2, and 8 opinion-unaware approaches viz., QAC, SISBLIM, NIQE, FISBLIM, CPBD, S3 and Fish_bb, were executed for the evaluation of the THz security image quality. The statistical results demonstrated the superiority of Fish_bb over the other testing IQA approaches for assessing the THz image quality with PLCC (SROCC) values of 0.8925 (-0.8706), and with RMSE value of 0.3993. The linear regression analysis and Bland-Altman plot further verified that the Fish__bb could substitute for the subjective IQA. Nonetheless, for the classification of THz security images, we tended to use S3 as a criterion for ranking THz security image grades because of the relatively low false positive rate in classifying bad THz image quality into acceptable category (24.69%). Interestingly, due to the specific property of THz image, the average pixel intensity gave the best performance than the above complicated IQA algorithms, with the PLCC, SROCC and RMSE of 0.9001, -0.8800 and 0.3857, respectively. This study will help the users such as researchers or security staffs to obtain the THz security images of good quality. Currently, our research group is attempting to make this research more comprehensive.

**Keywords**: terahertz security image quality assessment; THz image database; THz imaging technique; blind image quality assessment; THz security device

## Introduction

Recently, terahertz (THz) imaging technique is rapidly developing worldwide, spurred by its powerful capability of acquiring useful data in respect to physics, chemistry, biology and medicine [1, 2]. In contrast to the other imaging approaches, owing to the prominent merits of THz imaging technique such as low photon energy and high transparency [3], THz imaging technique has been extensively studied as an analysis tool in almost all basic and applied domains such as biological diagnosis [4] and security inspection [5]. Nonetheless, the current THz imaging devices require several seconds to collect one image, and therefore, the ultimate THz image quality is significantly influenced by the variations of environmental factors and the performances of equipment [6]. According to the previous researches, the present THz imaging systems always generate the THz images of sufficient low quality [7, 8], which in turn affects the detection or prediction accuracies and hinders the extension of THz imaging technique to the large scale applications. Hence, the assurance of THz image of good or acceptable quality is an indispensable procedure during the practical applications.

The use of objective image quality assessment (IQA) method can provide an efficient solution to quantitatively evaluate the THz image quality, thus allowing the subsequent improvement of the THz image quality in hardware or software [9]. IQA is extremely important for the numerous image and video processing tasks [10], aiming to automatically examine image quality in agreement with human quality judgments [11] or task requirements. The existing objective IQA approaches are in general classified into three categories, which are full-reference, reduced-reference and no-reference or blind methods, depending on the accessibility of reference images [12, 13]. Nowadays, IQA has been widely used in many multimedia applications inclusive of image compression and video transmission [14]. In terms of THz imaging technique, IQA can be used as a criterion for assessing the performance of imaging system, optimizing image acquisition and processing modules, and monitoring the working conditions of imaging components. For the reason that it is highly hard to capture the perfect reference THz images in the real world THz imaging, the no-reference IQA algorithms are recommended for the further analysis in this study.

For IQA of specific image modalities, Chow and Paramesran summarized the IQA algorithms for various medical image types containing magnetic resonance image, computed tomography and ultrasonic image [15]. Like natural images, the IQA algorithms for these specific image modalities can also be applied to optimize imaging protocol [16] and improve detection accuracy [17]. Very limited research work has been conducted on the design of IQA algorithms to assess THz image quality. Fitzgerald et al., calculated the modulation transfer function from the amplitude of optical transfer function to verify the spatial resolution of THz image [9]. Hou et al., computed the mean square error of the peak values of time domain in a column's pixel to evaluate THz image quality [6]. However, their researches are not comprehensive, and therefore, the further studies are urgently needed.

There are some publications using IQA for assessing security image quality in specific modalities. Long and Li leveraged five image features viz., sharpness, brightness, resolution, head pose and expression to evaluate the NIR face image quality [18]. Wu et al., developed IQA software for monitoring the performance of X-ray security screening system [19]. A similar research was reported in a research paper by Irvine et al., [20]. This team of investigators presented an IQA protocol to quantitatively evaluate the X-ray image quality for baggage screening. IQA algorithms were also applied as measures to identify whether the biometric samples were fake or real [21-23]. Nevertheless, to the best of our knowledge, there is no publication in respect to IQA of THz security image.

Thus, the objectives of the current work are to: (1) establish the THz security image database (THSID) and give the corresponding mean opinion scores (MOSs) for each image in proposed database; (2) adopt the existing no-reference IQA algorithms for the evaluation of the THz security images and automatically classify the THz security images into two grades; and (3) discuss the significance of IQA approach in the development of THz imaging technique and forecast the future research tendency.

## 2. Methods

### 2.1 THz security image database (THSID)

THSID contains a total of 181 THz security images with the resolution of 127×380. To generate THSID, four volunteers were invited to stand in the THz device, and imaged with various legal goods such as bracelet or illegal substances such as hammer each time. A raw THz image is a three-dimensional image cube, which can be revealed in Fig. 1. The two-dimensional THz image used in this paper is extracted from the three-dimensional THz image cube by combining the maximum pixels in the Z direct into the new two-dimensional plane.

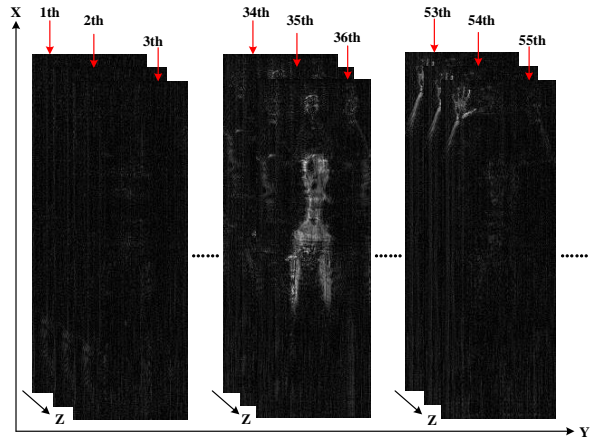

**Figure 1**. One sample for the explanation of three-dimensional image cube.

Some source THz security images are presented in Fig. 2.

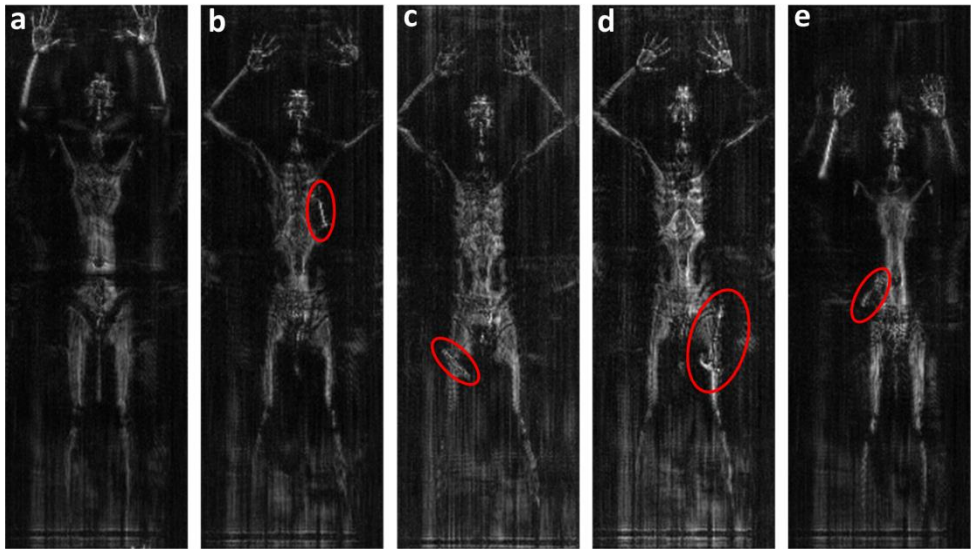

**Figure 2.** Some source THz security images in THSID: a) volunteer without any objects; b) volunteer carrying mineral water in right chest; c) volunteer taking knife in left pocket; d) volunteer carrying hammer in right pocket; and e) volunteer carrying plier in left abdomen. (Red circles are used to highlight the regions of legal or illegal substances)

### 2.2 Subjective test methodology

Before the subjective experiments, we analyzed main distortion types in THz security images. Fig. 3 shows one THz image with the volunteer taking phone in the right pocket, and the ripple noises caused by the unstable imaging device and changeable experimental factors make this target to be detected unidentified by human eyes or imaging processing algorithms.

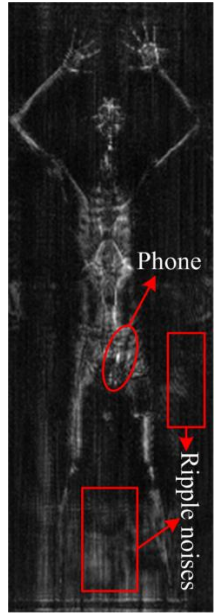

**Figure 3.** One THz image example for the analysis of main distortion type in THz security images. (Red circle and rectangles are applied for highlighting the illegal target and ripple noises respectively)

For THz images, from the above analysis, these global ripple noises are considered as the major factors to seriously deteriorate THz security image quality, which in turn decrease the detection accuracy of security inspection equipment. Consequently, the following subjective evaluation criteria (Table 1) are proposed for subjective experiments.

**Table 1**. Subjective evaluation criteria using five-grade scale for image quality assessment of THz security image.

| Score | Quality | Level of overall noise |
|-------|---------|------------------------|
| 5 | Excellent | Acceptable |
| 4 | Good | Unacceptable, but not annoying |
| 3 | Fair | Slightly annoying |
| 2 | Poor | Annoying |
| 1 | Bad | Very annoying |

With respect to each grade, the corresponding reference THz images (Fig. 4) are selected by two experts for conducting the double-stimulus method [24].

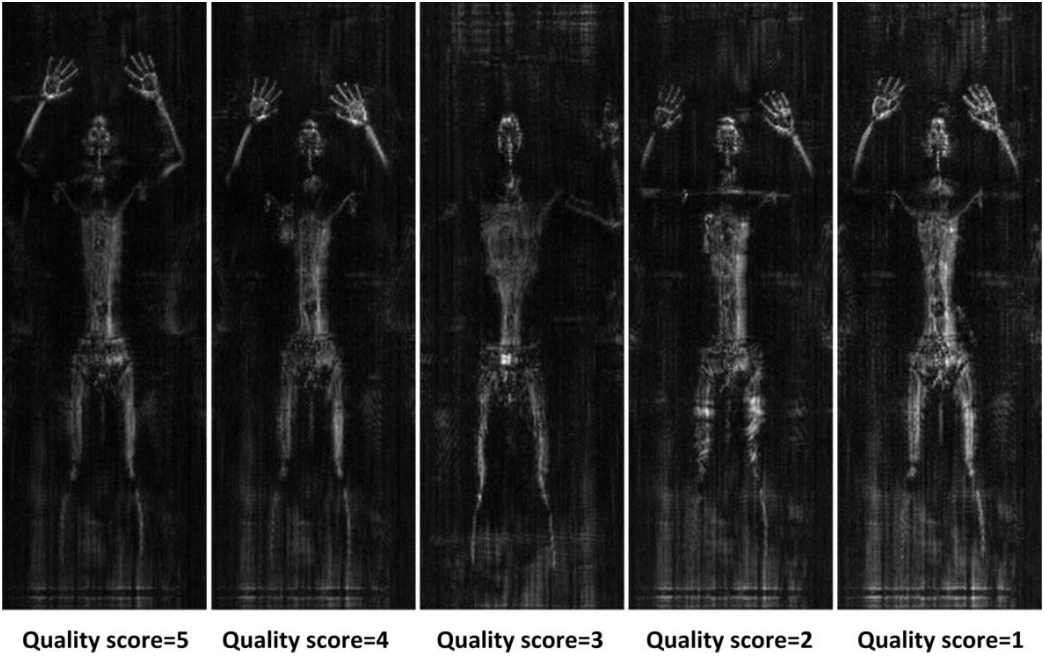

Quality score=5   Quality score=4   Quality score=3   Quality score=2   Quality score=1

**Figure 4.** Five reference THz security images for each grade presented in Table 1.

A total of 15 observers including 2 experts and 13 subjects having no related expertise were recruited to assess the quality of THz security images. Prior to tests, the objective and procedure of this experiment were individually introduced to all observers in detail. The experiments were carried out under the normal indoor illumination of about 2400 Lux. The testing images were displayed in the 23 inches LED monitor with the resolution of 1920×1080, and the viewing distance was set to 2-2.5 screen heights. After the experiments, the mean opinion scores (MOSs) of THz security images in THSID were obtained by the following equation.

$$MOS_j = \sum_{i=1}^{N} u_{i,j} / N_i \quad (1)$$

where $N_i$ and $u_{i,j}$ denote the number of observers and the score of image $j$ assigned by $i$th observer, respectively.

Fig. 5 demonstrates the statistical distribution of MOSs.

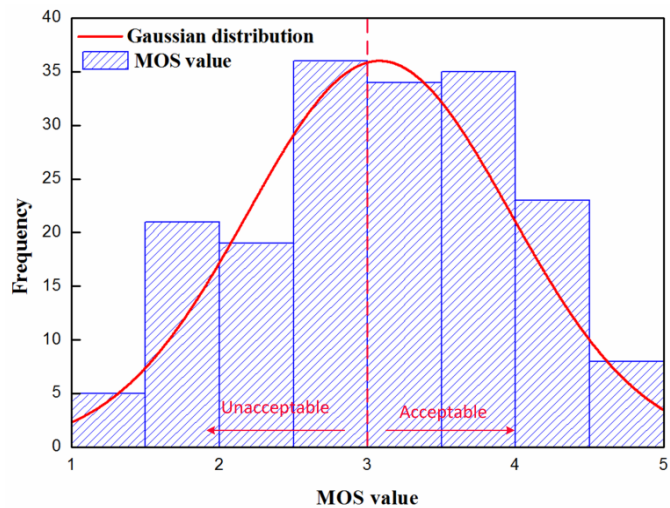

**Figure 5.** Histogram of MOS values for THz security images in THSID.

As shown in Fig. 5, the distribution of MOSs in THSID follows the Gaussian distribution, ensuring the comprehensive examination of the used IQA algorithms.

Furthermore, the THz images whose MOSs are below and up 3 are regarded as the bad and acceptable categories, respectively. Because we are more concerned with is that whether the THz image quality could meet the inspector's acceptability or final applicable requirements in practical applications. Subsequently, the k-means clustering was utilized to classify the THz images into two categories using IQA features.

### 2.3 No-reference IQA algorithms and performance metrics

A total of 13 no-reference IQA algorithms were executed in this study for evaluating the THz image quality. They are 5 opinion-aware approaches viz., NFERM [25], GMLF [26], DIIVINE [27], BRISQUE [28] and BLIINDS2 [29], and 8 opinion-unaware approaches viz., QAC [30], SISBLIM [31], NIQE [32], FISBLIM [33], CPBD [34], S3 [35] and Fish_bb [36]. Notice that the opinion-aware IQA methods used in this work employed natural images to establish the regression models.

In order to calculate the performance metrics, the monotonic logistic function of five parameters $\{\beta_1, \beta_2, \beta_3, \beta_4, \beta_5\}$ was first leveraged to fit the MOSs estimated by IQA algorithms, and the equation below was afterwards used to map the fitted objective scores to the subjective scores.

$$\text{Mapped Quality}(x) = \beta_1\left(1/2 - 1/\left(1 + e^{\beta_2(x-\beta_3)}\right)\right) + \beta_4 x + \beta_5 \qquad (2)$$

where Mapped Quality and $x$ are mapped objective score and its original score.

Subsequently, three metrics viz., Pearson Linear Correlation Coefficient (PLCC), Spearman Rank-Order Correlation Coefficient (SROCC) and Root Mean Squared Error (RMSE) were used to examine the performances of IQA algorithms on THz security images. Moreover, the linear correlation plot and the Bland-Altman plot [37] were also used to check the effectiveness of IQA algorithms.

## 3. Results and discussion

### 3.1 No-reference IQA for THz security images

The statistical data regarding the performances of five opinion-aware no-reference IQA methods is presented in Table 2. As shown in Table 2, BRISQUE was promising for the estimation of THz security images with the PLCC and SROCC as well as RMSE of 0.8001 and -0.7560 as well as 0.5309, respectively. The performances of DIIVINE and NFERM were slightly inferior to that of BRISQUE, with nearly 0.0277 (0.0312) and 0.0113 (0.0132) decrement (increment) in PLCC (RMSE), respectively. Both GMLF and BLIINDS2 did not yield desired results for the prediction of THz security images with the PLCC (RMSE) values below (beyond) 0.3500 (0.8000).

**Table 2.** Performance metrics of five opinion-aware blind IQA methods for THz security image database. (The superior results are highlighted in boldface)

| IQA \ Criteria | NFERM | GMLF | DIIVINE | BRISQUE | BLIINDS2 |
|---|---|---|---|---|---|
| | score | out | q | qualityscore | predicted_score |
| PLCC | 0.7888 | 0.3298 | 0.7724 | **0.8001** | 0.2929 |
| SROCC | 0.5539 | -0.2904 | -0.7271 | **-0.7560** | -0.2660 |
| RMSE | 0.5441 | 0.8356 | 0.5621 | **0.5309** | 0.8463 |

Table 3 summarizes performance metrics of seven opinion-unaware no-reference IQA algorithms for THSID database. As shown in Table 3, Fish_bb gave the best predictions than the other IQA approaches for assessing THz image quality, with PLCC (SROCC) values of 0.8925 (-0.8706), and with RMSE value of 0.3993. The performances of QAC, S3 and SISBLIM were comparable with that of Fish_bb, which produced the PLCC (SROCC) values of 0.8280 (0.7985), 0.8716 (-0.8560) and 0.8186 (-0.7915), and with RMSE values of 0.4963, 0.4339 and 0.5083, respectively. NIQE, FISBLIM, CPBD and SINE resulted in low PLCC (SROCC) and high RMSE values for the assessment of THz security image quality (below 0.5600 (0.4600) and beyond 0.7300, respectively). In contrast, Fish_bb achieved the slightly superior overall performance to S3, with the PLCC (SROCC) values of 0.8925 (-0.8706) versus 0.8716 (-0.8560), and with RMSE value of 0.3993 versus 0.4339 (Table 3).

**Table 3.** Performance metrics of eight opinion-unaware blind IQA methods for THz security image database. (The superior results are highlighted in boldface)

| IQA Criteria | QAC | SISBLIM | NIQE | FISBLIM | CPBD | SINE | S3 | Fish_bb |
|---|---|---|---|---|---|---|---|---|
| | q | score3 | quality | ss | metric_cpbd | noise_SD | s31 | sh1 |
| PLCC | **0.8280** | **0.8186** | 0.4451 | 0.5512 | 0.2052 | 0.3457 | **0.8716** | **0.8925** |
| SROCC | **0.7985** | **-0.7915** | -0.4025 | -0.4501 | 0.1122 | -0.3907 | **-0.8560** | **-0.8706** |
| RMSE | **0.4963** | **0.5083** | 0.7926 | 0.7385 | 0.8663 | 0.8305 | **0.4339** | **0.3993** |

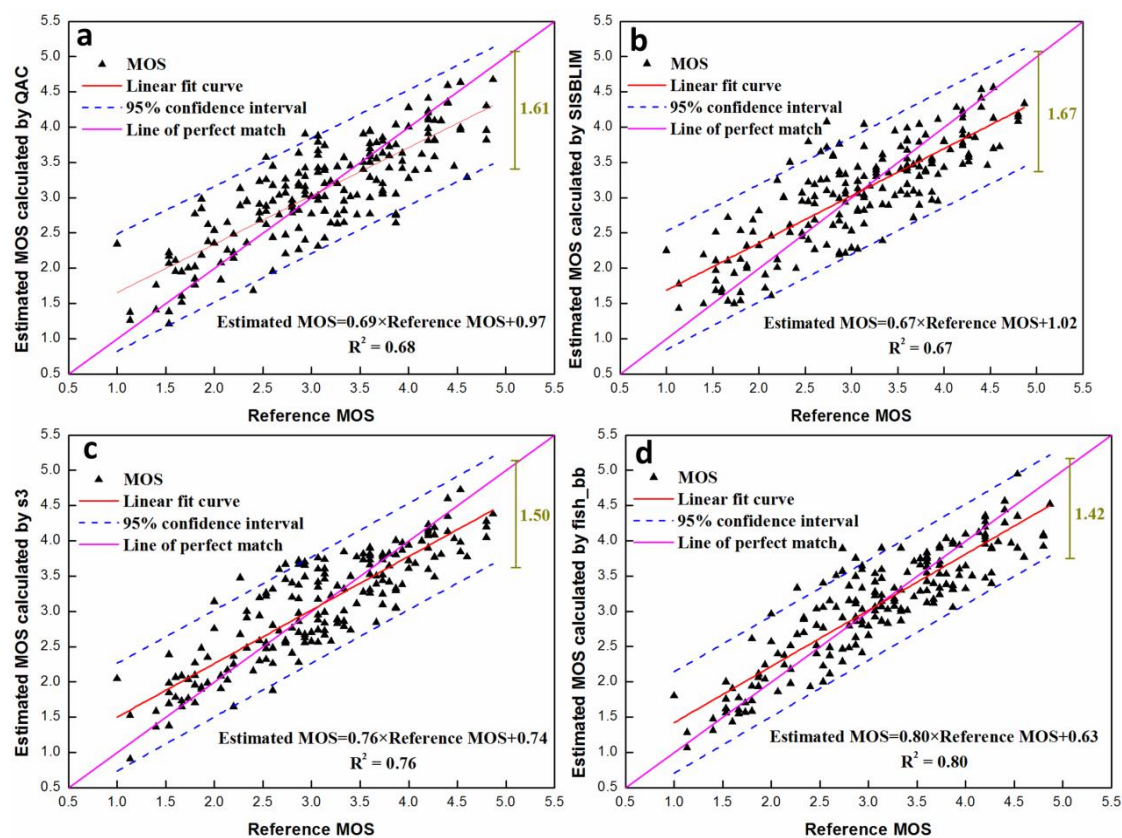

**Figure 6.** Scatter plots and linear regressions of reference MOSs versus predicted quality using QAC (a), SISBLIM (b), S3 (c) and Fish_bb (d).

Fig. 6 visualizes the distributions of the estimated and reference MOSs. As shown in Fig. 6, for QAC, SISBLIM, S3 and Fish_bb, the scatter plots of predicted against reference MOSs clearly revealed that a majority of the THz image samples were mainly close to the line of perfect match (slope = 1) and within the 95% confidence intervals. The vertical distances between the upper and lower 95% confidence intervals were 1.61, 1.67, 1.50 and 1.42 for QAC, SISBLIM, S3 and Fish_bb respectively, indicating that Fish_bb performed better than the other algorithms for THz images in THSID database. In the case of linear regression analysis, Fish_bb was further verified to outperform QAC, SISBLIM and S3 with determination coefficients ($R^2$) of 0.80 versus 0.68, 0.67 and 0.76.

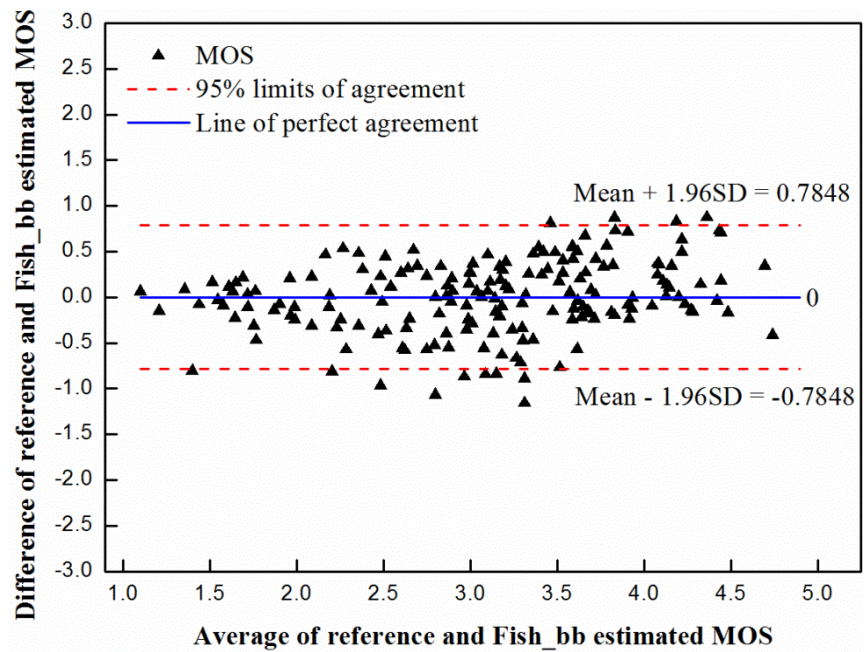

**Figure 7.** Bland-Altman plot ($N = 181$ image samples) of the difference against average for THz IQA using subjective evaluation and Fish_bb objective IQA algorithm.

The Bland-Altman plot of reference and Fish_bb estimated MOS is demonstrated in Fig. 7. It could be observed that the distribution of points in Bland-Altman plot in Fig. 7 was quite similar to that in Fig. 6(d). Approximately 93.9% of data points located within the limits of agreement of 0.7848 and -0.7848 and the majority of points were dispersed around the line of perfect agreement, which further indicated that the Fish_bb objective IQA algorithm could substitute for the subjective IQA.

### 3.2 Classification of THz security images

In real world applications, we are more concerned with is that whether the THz image quality could meet the inspector's acceptability or final applicable requirements. Table 4 summarizes the unsupervised classification results of THz security image quality based on four IQA algorithms. The four IQA features viz., q, score3, s31 and sh1 achieved the overall classification accuracies of 77.35%, 78.45%, 79.56% and 83.98%, respectively. Although sh1 estimated by Fish_bb yielded the best classification result, 26 THz images of bad quality were classified as images of acceptable quality (Table 4). This illustrated that 26 THz images of bad quality would display on the screen for inspectors when 81 images of bad quality were generated by THz device. In this sense, we preferred to choose s31 calculated by S3 as a criterion to screen the THz images of bad quality with relatively low false positive rate of 24.69% (Table 4).

**Table 4.** Unsupervised classification results of THz security image quality using four IQA features.

| IQA feature | Quality category | No. of sample | Classification result | | Accuracy/% | Overall accuracy/% |
|---|---|---|---|---|---|---|
| | | | Acceptable | Bad | | |
| q | Acceptable | 100 | 85 | 15 | 85.00 | 77.35 |
| | Bad | 81 | 26 | 55 | 67.90 | |
| score3 | Acceptable | 100 | 91 | 9 | 91.00 | 78.45 |
| | Bad | 81 | 30 | 51 | 62.96 | |
| s31 | Acceptable | 100 | 83 | 17 | 87.00 | 79.56 |
| | Bad | 81 | 20 | 61 | 75.31 | |
| sh1 | Acceptable | 100 | 97 | 3 | 97.00 | 83.98 |
| | Bad | 81 | 26 | 55 | 67.90 | |

### 3.3 Discussion

From Table 2 and Table 3, it could be observed that many general-purpose no-reference IQA algorithms worked not so well on the constructed THz security image database. It was not surprising since those algorithms were implicitly designed for natural images, especially these natural scene statistics (NSS) based methods, while THz security images largely deviated from NSS. From Table 3, we could observe that two general-purpose no-reference IQA algorithms viz., QAC and SISBLIM worked relatively well. QAC performed quality-aware clustering, in which patches of the same quality level were clustered. During the training procedure of QAC, noisy images were used, and thus there could be clusters of noisy patches. Noisy patterns of natural images and THz images were very similar. This might be the reason why QAC performed relatively well. Similarly, SISBLIM also contained a module estimating noise. Besides these two algorithms, two sharpness algorithms viz., S3 and Fish_bb also worked relatively well. As illustrated in Fig. 4, it can be observed that high-quality THz security images have clean background, and the whole image is relatively smooth except for the target. Nonetheless, in THz images of low quality, the whole image is quite noisy and sharp. Consequently, sharpness algorithms can have low quality scores for THz images of high quality and high quality scores for THz images of low quality (data not shown), which is contrary to natural images. This reversed phenomenon has no effect on the final predictive ability of sharpness algorithm.

We tried to develop a new IQA algorithm to evaluate the THz image quality; however, the average pixel intensity of THz image achieved the good performance with the PLCC, SROCC and RMSE of 0.9001, -0.8800 and 0.3857, respectively, outperforming the Fish_bb and S3. The possible reason for this might be due to the fact that the THz images were to some extent similar to the binary images. The subjects perceived the white pixels in THz images as the noises, thus making the average pixel intensity work well. Hence, in terms of assessing noise level in THz images, the average pixel intensity could be used instead of the complicated IQA algorithms. In practice, the new subjective evaluation criteria were urgently required to obtain MOS values that could reflect both the noise level and perceived quality in region of interest (e.g. the regions containing illegal substances).

Actually, we were not sure that whether the THz display mode used in this work was better than the other display modes or not. For example, we could also use the cross profile of maximum average pixel intensity in Z direction to be displayed in the viewing screen for security staffs. Therefore, there existed two possible research contents for the further study: the first is to devise a rule (this rule could be an IQA algorithm) to confirm one best THz display mode for commercial applications and scientific researches, and the second is to design a series of IQA approaches for subjective and objective quality assessment of THz image (these algorithms could be developed on the basis of three-dimensional or two-dimensional THz images).

The possible application perspectives of the IQA of THz image quality could refer to commercial applications and scientific researches. For commercial applications, as shown in Fig. 8, the IQA indicator can guarantee THz security images of good quality displaying on the screen via reshoot or shooting parameters reset. With respect to scientific researches, we can use the IQA indicator as a criterion to assist in developing new THz security image processing algorithms such as THz image enhancement for THz imaging device (Fig. 8).

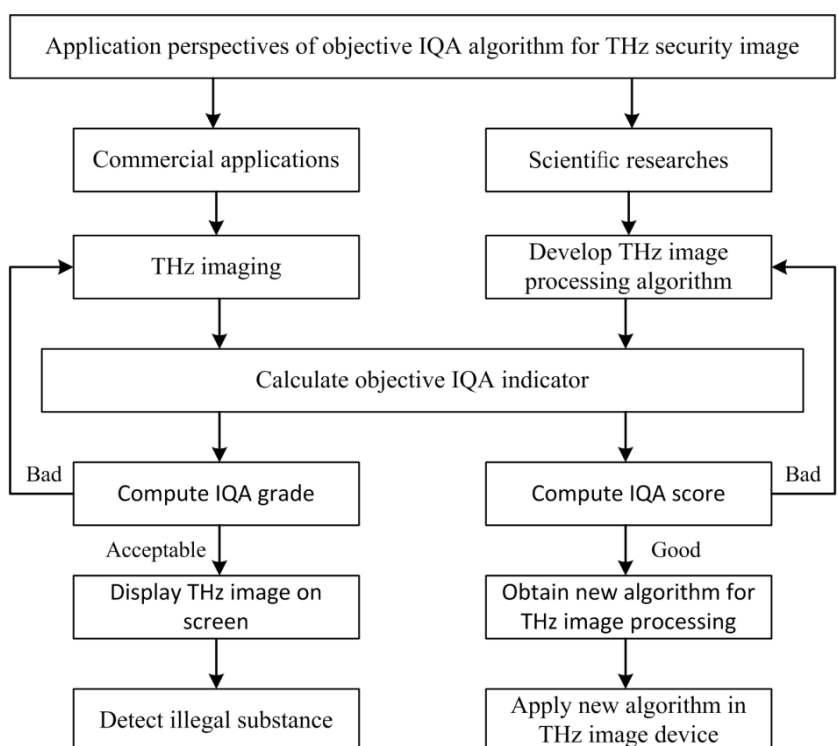

**Figure 8.** Role of THz IQA algorithms in the overall THz security imaging system.

## 4. CONCLUSION

In this paper, we constructed the THz security image database (THSID) inclusive of 181 THz security images with the resolution of 127×380. The results of statistical metrics demonstrated that the Fish__bb objective IQA algorithm outperformed the existing no-reference IQA algorithms viz., NFERM, GMLF, DIIVINE, BRISQUE, BLIINDS2, QAC, SISBLIM, NIQE, FISBLIM, CPBD and S3, with PLCC (SROCC) values of 0.8925 (-0.8706), and with RMSE value of 0.3993. The linear regression analysis and Bland-Altman plot further verified that the Fish__bb could substitute for the subjective IQA. In respect to the classification of THz security images, S3 was preferred to be applied as an indicator for screening THz security image grades because of the low classification rate in classifying bad THz image quality into acceptable category. However, we found that the average pixel intensity could give the best performance than the above complicated

IQA algorithms. Therefore, the average pixel intensity could be used as the indicator to evaluate the noise level of THz image in the practical application.

## Acknowledgements

The authors would like to acknowledge the staffs working in BOCOM Smart Network Technologies Inc., who assisted in acquiring the THz images.